\documentclass[10pt,twocolumn]{article} 
\usepackage{simpleConference}
\usepackage{times}
\usepackage{graphicx}
\usepackage{amssymb}
\usepackage{url,hyperref}

\usepackage{amsmath}
\usepackage{authblk}
\usepackage{algorithm,algorithmic}
\usepackage{float}
\usepackage{subcaption} 
\usepackage{tabularx}

\begin{document}

\title{Semi Few-Shot Attribute Translation}

\author{Ricard Durall$^{1,2,3}$ \qquad
Franz-Josef Pfreundt$^{1}$ \qquad
Janis Keuper$^{1,4}$\\
$^1$Fraunhofer ITWM, Germany\\
$^2$IWR, University of Heidelberg, Germany\\
$^3$Fraunhofer Center Machine Learning, Germany\\
$^4$Institute for Machine Learning and Analytics, Offenburg University, Germany\\
}

\maketitle
\begin{abstract}
Recent studies have shown remarkable success in image-to-image translation for attribute transfer applications. However, most of existing approaches are based on deep learning and require an abundant amount of labeled data to produce good results, therefore limiting their applicability. In the same vein, recent advances	in meta-learning have led to successful implementations with limited available data, allowing so-called \textit{few-shot learning}.

In this paper, we address this limitation of supervised methods, by proposing a novel approach based on GANs. These are trained in a meta-training manner, which allows them to perform image-to-image translations using just a few labeled samples from a new target class. This work empirically demonstrates the potential of training a GAN for few shot image-to-image translation on hair 	color attribute synthesis tasks, opening the door to further research on generative transfer learning.
\end{abstract}

\section{Introduction}

Deep learning models have achieved state-of-the-art performance in large varity of tasks, from image synthesis \cite{zhang2017stackgan,karras2017progressive,bao2017cvae}, text style transfer \cite{shen2017style}, video generation \cite{bansal2018recycle, wang2018video} to image-to-image translation \cite{pathak2016context,iizuka2017globally,zhu2017toward,choi2018stargan,mo2018instanceaware}. The latter task, image-to-image translation, is a computer vision problem that aims at translating images from one domain to another, including colorization \cite{zhang2016colorful}, super-resolution \cite{ledig2017photo}, style transfer \cite{isola2017image,zhu2017unpaired,zhu2017toward,huang2018multimodal}, inpainting \cite{pathak2016context,yeh2017semantic,li2017generative,iizuka2017globally,yu2018generative} and attribute transfer \cite{liao2017visual,kim2017learning,creswell2017adversarial,choi2018stargan}. The strong performance, however, heavily relies on training a network with abundant labeled instances with diverse visual variations. The human annotation cost as well as the scarcity of data in some classes significantly limit the applicability of current vision systems to learn new visual concepts efficiently. In contrast, the human visual systems can recognize new instances with extremely few labeled examples. It is thus of great interest to learn to generalize to new cases with a limited amount of labeled examples for each novel case.

The problem of learning to generalize to unseen classes during training, known as few-shot classification, has attracted considerable attention \cite{vinyals2016matching,ravi2016optimization,finn2017model,snell2017prototypical,sung2018learning}. One promising direction to few-shot classification is the meta-learning paradigm where transferable knowledge is extracted and propagated from a collection of tasks to prevent overfitting and improve generalization. Recent advances in meta-learning algorithm includes metric-based methods \cite{koch2015siamese,vinyals2016matching,snell2017prototypical,sung2018learning}, model-based methods \cite{santoro2016meta,munkhdalai2017meta} and optimization-based methods \cite{ravi2016optimization,finn2017model,antoniou2018train,nichol2018first,finn2018probabilistic}. These models have allowed learning tasks to perform well on novel data sampled from the same distribution as the training data. These meta-learning algorithms have seen direct applications in supervised and reinforcement learning. Additionally, due to their general applicability, recent works based on meta-learning have successfully been utilized for image generation \cite{lake2011one,rezende2016one,bartunov2018few,clouatre2019figr}.

The objective of attribute transfer is to synthesize realistic appearing images for a pre-defined target domain. For instance, given an image with a particular attribute "blond hair" (original domain), change it to "black hair" (target domain). We refer to a domain as a set of images sharing the same attributes. Such attributes are meaningful semantic feature inherent in an image such as "smiling" or "face with eyeglasses". After the introduction of generative adversarial networks (GANs) \cite{goodfellow2014generative}, transfer domain algorithms have experienced significant improvements achieving state-of-the-art results in style transfer \cite{isola2017image,zhu2017unpaired,zhu2017toward,huang2018multimodal} and in attribute transfer \cite{kim2017learning,creswell2017adversarial,choi2018stargan}. However, GANs require several orders of magnitude more data points than humans in order to generate comprehensible images successfully from a given class of images \cite{clouatre2019figr}. This impairs the ability of GANs to generate novelty. Additionally, in many cases, if the data is abundant enough to successfully train a GAN, there is little purpose to generating more of this data. 

In this work we focus on the challenging scenario, where we define the problem of semi few-shot image-to-image translation. In particular, we propose a novel approach capable of performing attribute transfer on a target domain with a very limited amount of labeled data. In order to achieve this goal, we use two independent but equal networks and we train them as proposed in \cite{nichol2018first}. Most of the existing few-shot algorithm are applied to classification tasks where one class remains unseen. In our approach, we apply the same principle but with attributes and we name it semi few-shot because there are not different classes. Nevertheless, having only one kind of images but with different attributes does not make the problem trivial since still the network needs to learn the ability to perform image-to-image transformation for untrained target domain with a few examples. We apply our model to the CelebA \cite{liu2015deep} dataset of faces and control several hair color attributes.

Overall, our contributions are summarized as follows

\begin{itemize}
\item  We propose a novel generative adversarial network based on meta-learning, trained for end-to-end image-to-image translation.

\item We demonstrate, how we can successfully learn to transfer hair attributes by using a generative few-shot approach. These first results are opening the door to further research.

\item We provide qualitative results based on CelebA dataset, showing the effectiveness of our proposal.
\end{itemize}

\section{Related Work}
While machine learning systems might have surpassed humans at many tasks \cite{dodge2017study}, they generally need far more data to reach the same level of performance. Nonetheless, it is not completely fair to compare humans to algorithms directly, since humans enter a task with a large amount of prior knowledge. In other words, humans do not learn from scratch, but they fine-tune and recombine sets of pre-existing skills. Meta-learning has emerged recently as an approach for learning from small amounts of data as human do. This machine learning field has already been employed in many different domains such as classification, reinforcement learning and even image generation.

\subsection{Meta-learning}

Meta-learning, also known as few-shot learning, intends to design models that can learn new skills or adapt to new environments rapidly with a few training examples. Specifically, we assume to have access to a problem, which is split into a set of tasks. Each of these tasks can be for example: a set of images (belonging to the same class) for a classification problem, or a set of state, action and reward for a reinforcement algorithm, or even an attribute for an image-to-image transformation. Then, from this problem, we sample a training set and a test set of tasks, and we feed the training set into our algorithm, so that, eventually it will produce good performance on the test set. Since each task corresponds to a learning problem, performing well on a task corresponds to learning quickly.

A variety of different approaches to meta-learning have been proposed, each with its own flavors. There are three common approaches: metric-based that learns an efficient distance metric \cite{koch2015siamese,vinyals2016matching,snell2017prototypical,sung2018learning}, model-based that uses network with external or internal memory \cite{santoro2016meta,munkhdalai2017meta} and optimization-based that optimizes the model parameters explicitly for fast learning \cite{ravi2016optimization,finn2017model,antoniou2018train,nichol2018first,finn2018probabilistic}.

\vspace{2mm}\noindent \textit{Optimization-based.}
Optimization-based models, also known as initialization-based  methods, tackle the  meta-learning  problem  by  "learning  to  fine-tune". The idea behind this approach is to learn a good model initialization (i.e. the weights) so that in the case of a classification task, given an unseen class, the model can classify correctly using only a limited number of labeled examples and a small number of gradient update steps \cite{ravi2016optimization,finn2017model,antoniou2018train,nichol2018first,finn2018probabilistic}. These initialization based methods are capable of achieving fast and effective adaption with a limited number of training examples for new classes, however they still have difficulty in handling domain shifts between base and novel classes.

\vspace{2mm}\noindent \textit{Few-Shot Image Generation.}
Most of meta-learning applications focus on classification tasks defined as the ability to learn a classifier to recognize unseen classes during training with limited labeled examples. However, it is possible to extend such a definition of few-shot learning to other domains such as image generation. To best of our knowledge, \cite{lake2011one} pioneered the successful combination of few-shot techniques with image generation. In this first approach, the model is fed with images and their strokes, and trained on the Omniglot dataset \cite{lake2011one}. This yields a system that can generate novel binary samples. Trying to make a more general approach, \cite{rezende2016one} presents a sequential generative model which  is only trained on pure images (no stroke information is required). Even though, this second approach improves previous results, it suffers from lengthy sequential inference as a consequence. \cite{bartunov2018few} tackles this issue by suggesting matching networks (memory-assisted networks). This implementation generates binary images on	Omniglot dataset using few-shot learning and with fast inference periods.	Finally, \cite{clouatre2019figr} proposes a new approach  integrating GANs in the structure, surpassing in this way, the scalability limitation found in the other models. Furthermore, this work presents more extensive experiments including additional datasets (MNIST \cite{lecun2010mnist} and FIGR-8	\cite{clouatre2019figr}).

\subsection{Generative Adversarial Network}

Generative adversarial network \cite{goodfellow2014generative} is capable of learning deep generative models. It can be described as a minmax game between the generator $G$, which learns how to generate samples which resemble real data, and a discriminator $D$, which learns to discriminate between real and fake data. Throughout this process, $G$ indirectly learns how to model the input image distribution $p_{data}$ by taking samples $z$ from a fixed	distribution $p_{z}$ (e.g. Gaussian)  and forcing the generated samples $G(z)$ to match the natural images $x$. The objective loss function is defined as

\begin{align}
\begin{split}
	\min_{G} \max_{D} \mathcal{L}(D,G) =\,& \mathbb{E}_{\mathrm{\mathbf{x}} \sim p_{\mathrm{data}}} \left[ \log \left(D(\mathrm{\mathbf{x}})\right) \right] \,+\\	
	+\,& \mathbb{E}_{\mathrm{\mathbf{z}} \sim p_{\mathrm{z}}}[\log(1-D(G(\mathrm{\mathbf{z}})))].
\end{split}
\end{align}

\vspace{2mm}\noindent\textit{Domain Transfer.}
GAN-based approaches for image-to-image translation have been actively studied. One of the first proposals \cite{isola2017image} capable of learning consistent image domain transforms, employed a pair of images that could be used to create models that convert input from an original domain to different target domain (e.g. segmentation labels to the original image). But this system requires that both images and target images must exist as pairs in the training dataset in order to learn the transformation between domains. Several works \cite{zhu2017unpaired,choi2018stargan} try to address this drawback. Their suggestion is to use the virtual result in the target domain. Therefore, if the virtual result is inverted again, the inverted result must match with the original image.  In these works, the framework can control the image translation into different target domains. Recently, numerous works have focused on transferring visual attributes such as color \cite{zhang2016colorful}, texture \cite{isola2017image,zhu2017unpaired,zhu2017toward,huang2018multimodal}, facial
features \cite{liao2017visual,kim2017learning,creswell2017adversarial,choi2018stargan} and more.  However, although most of these approaches synthesize new images that belong to the target domain, they lack in generalizing attributes since they are designed to transfer a specific type of visual attribute.

\section{Method}
In the following section, we describe our approach which addresses image-to-image translation, where the input image contains the original attributes and the output image the target attributes. We explain the training of the model in two levels of abstraction that train independently, following an adversarial fashion which allows generating realistic samples containing the target attributes.

\begin{figure}[t]
\begin{center}
   \includegraphics[width=1\linewidth]{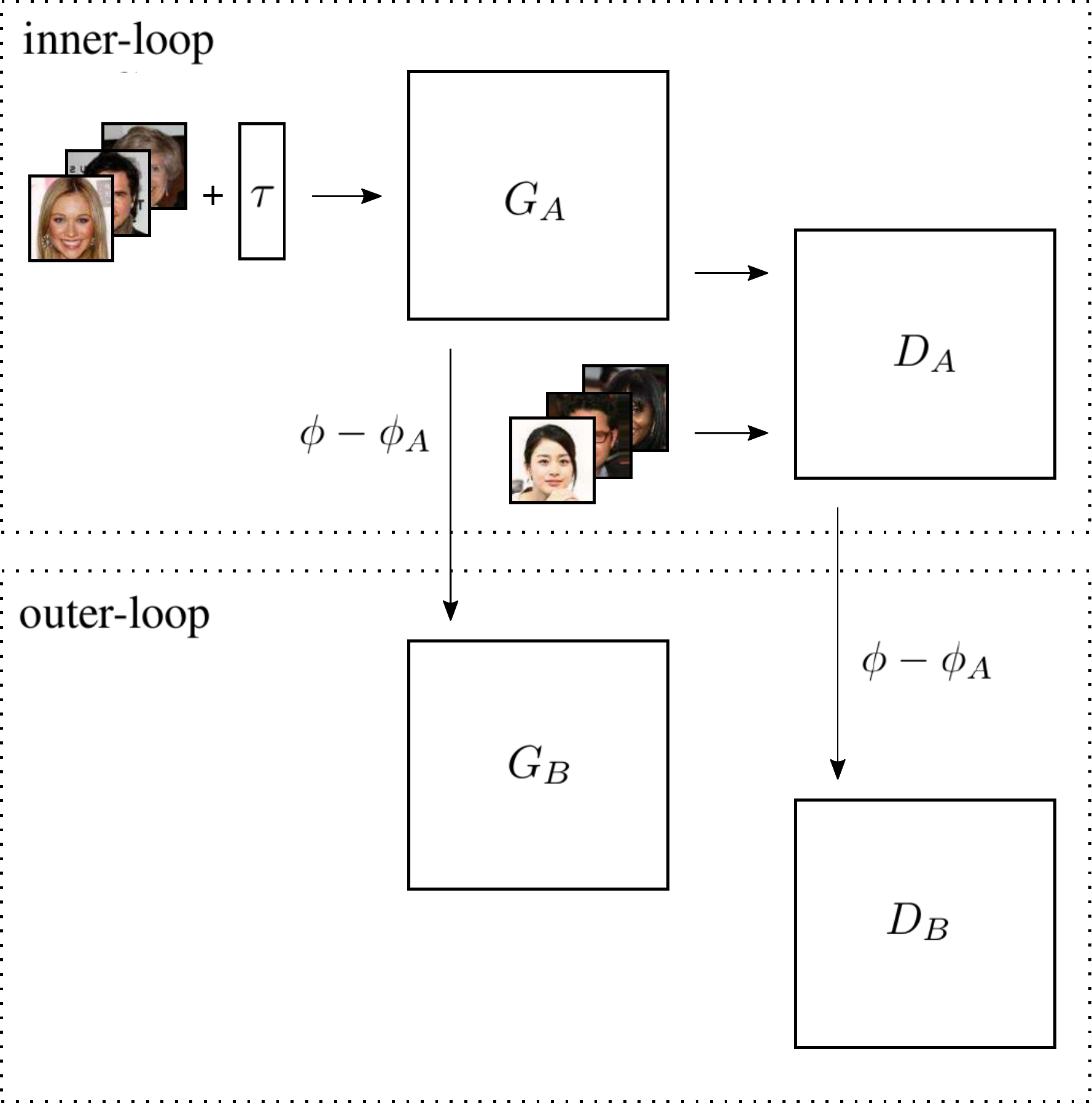}
\end{center}
   \caption{The figure shows the few-shot attribute transfer structure, which consists of two distinguishable parts: inner-loop and outer-loop. At the same time, each of this blocks is a GAN system formed for a generator $G$ and a discriminator $D$. The inner-loop optimizes $G_{A}$ using samples $A$ and the task $\tau$ judged by $D_{A}$. After $k$ updates, we update the outer-loop by setting its gradient to $\phi - \phi_A$ and performing one step of optimizer.}
   \label{fig:1}
\end{figure}

\begin{figure*}
\begin{center}
   \includegraphics[width=0.8\linewidth]{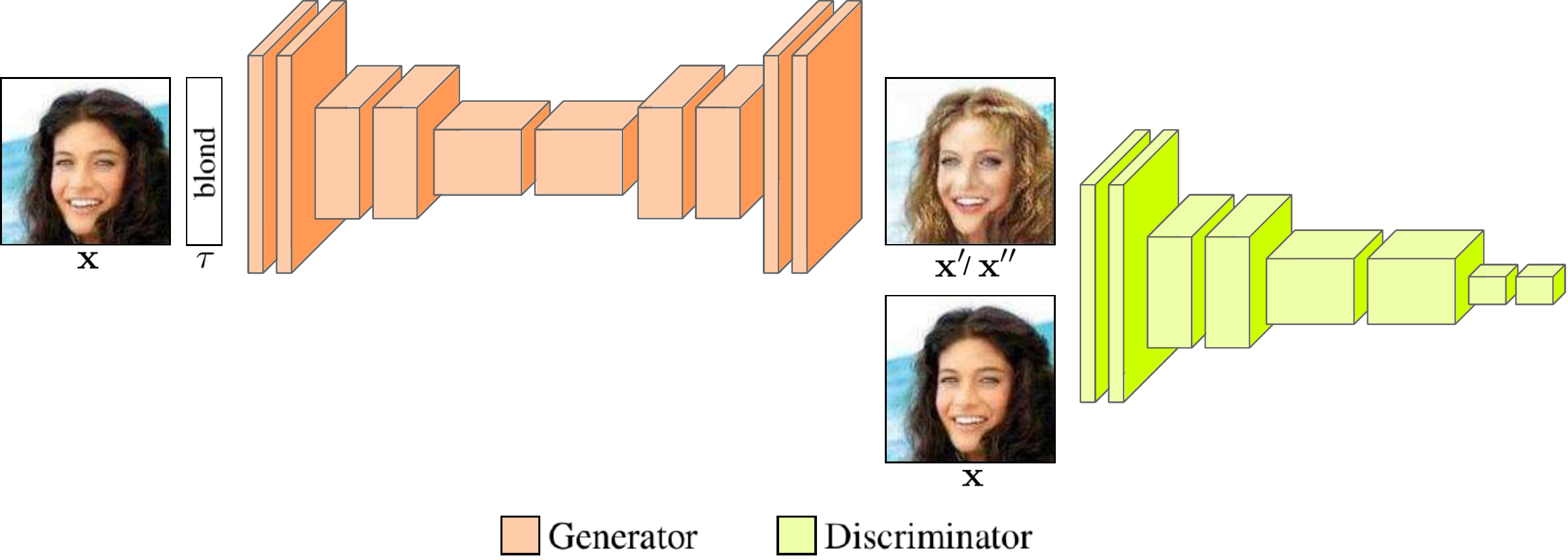}
\end{center}
   \caption{Overview of the attribute transfer network structure while training. Our model contains a generator $G$ and a discriminator $D$, which are trained independently. For example, given an input image and a target domain label (blond), $G$ learns to generate images within the target domain. At the same time, $D$ learns to distinguish between real and generated images, and to classify them to their corresponding domain.}
\label{fig:2}
\end{figure*}

\subsection{Meta-learning Model Architecture}
We define the few-shot attribute transfer problem $P(\mathcal{T})$ based on the meta-learning reptile algorithm introduced in \cite{clouatre2019figr}. In this scenario, we assume a set of tasks $\{\mathcal{T}_{i}\}_{i=1}^{M}$, where each individual task $\tau$ is an attribute transfer problem with its corresponding loss $L_{\tau}$. The intuition behind $L_{\tau}$ is that it accounts for the ability of generating realistic samples. We approach our problem by defining a meta-learning model that optimizes towards convergence, by modifying the neural
network parameters $\phi$ within a limited $k$ updates. Therefore, we can define the minimization problem as

\begin{align}
\begin{split}
	\min_{\phi} \mathbb{E}_{\tau}[L_{\tau}(U_{\tau}^{k}(\phi))],
\end{split}
\label{eq:1}
\end{align}

\vspace{2mm}where $U_{\tau}^{k}(\phi)$ is the operator (usually stochastic gradient descent) that updates $\phi$ parameters $k$ times using data sampled from $\tau$. 

In order to solve Equation \ref{eq:1}, we employ a solution composed of two independent neural structures, but with the same topology, called outer-loop structure and an inner-loop structure. Given a task $\tau$ and the initial parameters $\phi$, the inner-loop optimization trains on a set of samples, while updating a copy of the parameters $\phi_A$. After $k$ updates, we optimize the outer-loop, where we set the gradient of $\phi_B$ to be equal to $\phi - \phi_A$, take one step on the optimizer and compute the loss. In this way, the algorithm optimizes for generalization. We write it as

\begin{align}
\begin{split}
	\min_{\phi} \mathbb{E}_{\tau}[L_{\tau}(U_{\tau,A}^{k}(\phi_A))].
\end{split}
\label{eq:2}
\end{align}

\vspace{2mm}Notice that this dual structure gives enough flexibility to cope with different problems using the same framework. In this work, we make use of such a generalization feature by implementing a GAN system inside of each loop (see Figure \ref{fig:1}). For a given image $\mathrm{\mathbf{x}}$ and a target domain label $\tau$, our goal is to translate $\mathrm{\mathbf{x}}$ into an output image $y$, which now belongs to the target domain. In other words, we can transfer attributes by generating synthetic data in a meta-learning
fashion.

\subsection{Attribute Transfer Model Architecture}
We construct our baseline attribute transfer generative network based on recent state-of-the-art image-to-image translation model \cite{choi2018stargan} which is an adaptation from \cite{zhu2017unpaired}.  In particular, our approach employs an adversarial structure made of a generator and a discriminator. By combining these elements, our model can self-estimate the difference between the generated samples (with attribute transfer) and the real samples, and then update itself to produce more realistic samples. The network architecture of
our proposal is shown in Figure \ref{fig:2}.

\vspace{2mm}\noindent \textit{Generator.}
The goal of the generator $G$ is to learn mappings among multiple attribute domains. To achieve this goal,  we train $G$ to translate $\mathrm{\mathbf{x}}$ into an output image $\mathrm{\mathbf{x'}}$ conditioned on the target domain label $\tau$. To achieve this objective, the generator loss consists of $\mathcal{L}_{\mathrm{disc}}$ that penalizes inappropriately generated images and $\mathcal{L}_{cycle}$ that guarantees that translated images $\mathrm{\mathbf{x'}}$ preserve the content of its input images $\mathrm{\mathbf{x}}$  while changing only the domain-related part of the inputs. We write the generator loss as

\begin{align}
\begin{split}
	\mathcal{L}_{\mathrm{gen}} =  \,&  \mathcal{L}_{\mathrm{disc}}  \,+\, \lambda_{\mathrm{cycle}} \,\mathcal{L}_{\mathrm{cycle}}.
\end{split}
\end{align}

\vspace{2mm}Notice that the generator performs an entire cyclic translation $\mathrm{\mathbf{x}}\rightarrow \mathrm{\mathbf{x'}} \rightarrow \mathrm{\mathbf{x''}}$ for every sample, forcing $\tau$ to be crucial for moving among domains. First, to translate an original image $\mathrm{\mathbf{x}}$ into an image in the target domain $\mathrm{\mathbf{x'}}$
and then to reconstruct the original image from the translated image $\mathrm{\mathbf{x''}}$. This procedure can be written as

\begin{align}
\begin{split}
	 \mathcal{L}_{cycle} = \,& ||\mathrm{\mathbf{x}} \,-\, \mathrm{\mathbf{x''}} ||_{1}
\end{split}
\end{align}

\vspace{2mm}where 

\begin{align}
\begin{split}
	\mathrm{\mathbf{x'}} = \,& G(\mathrm{\mathbf{x}}, \tau_{\mathrm{target}}) \\ \mathrm{\mathbf{x''}} = \,& G(\mathrm{\mathbf{x'}}, \tau_{\mathrm{orginal}}).
\end{split}
\end{align}

\vspace{2mm}\noindent \textit{Discriminator.}
The second element of the model is the discriminator $D$. It takes samples of true and generated data and tries to classify them correctly. This classification procedure takes place after reconstruction and generation tasks, and it consists of two parts. One part that implements $\mathcal{L}_{\mathrm{adv}}$, which employs Wasserstein distance to determine
if an image looks realistic and penalize it otherwise, and a second part similarly to other approaches like \cite{chongxuan2017triple}, where we have an additional loss function $\mathcal{L}_{\mathrm{class}}$ which accounts for domain classification. This term computes the binary cross entropy loss between $\tau_{\mathrm{generated}}$ and $\tau_{\mathrm{target}}$ penalizing incorrect image-domain transformations. Therefore, the discriminator loss can be defined as

\begin{align}
\begin{split}
	\mathcal{L}_{\mathrm{disc}} =  \,& \lambda_{\mathrm{adv}} \, \mathcal{L}_{\mathrm{adv}}  \,+\, \lambda_{\mathrm{class}} \, \mathcal{L}_{\mathrm{class}}.
\end{split}
\end{align}

\section{Experiments}

In this section, we present experimental results evaluating the effectiveness of the proposed method. First, we give a detailed introduction of the experimental setup. Then, we discuss the results and its possible interpretation.

\subsection{Experimental Settings}
We train our model on the CelebFaces Attributes (CelebA) dataset \cite{liu2015deep}. It consists of 202,599 celebrity face images with variations in facial attributes. For the experiments, since we focus on hair attributes, we select a set of 32,502 images containing a balanced amount of hair attributes (blond, black, brown and gray hair). It is indispensable to have such an even distribution, otherwise the algorithm might fail at transferring marginal attributes. We randomly select 2,000 images for testing and use all remaining images as the training dataset. In training, we crop and resize the initially 178x218 pixel image to 128x128 pixels. All experiments presented in this paper have been conducted on a single NVIDIA GeForce GTX 1080 GPU.

\subsection{Training Setting}

Since our model is divided into two distinguishable parts, two independent Adam optimizer \cite{kingma2014adam} with $\beta_{1} = 0.5$, $\beta_{2} = 0.999$. are used during training. We set the batch size to 16 and run the experiments for 200K iterations. We update the generator after every five discriminator updates as in \cite{gulrajani2017improved, choi2018stargan}. The learning rate used in the implementation is 0.0001 for the first 10 epochs and  then linearly decreased to 0 over the next 10 epochs. The losses are weighted by the factors:
$\lambda_{\mathrm{class}}$ and  $\lambda_{\mathrm{cycle}}$ set to 10, and $\lambda_{\mathrm{adv}}$ to 1. The inner-loop $k_{A}$ runs 1 iteration and the outer-loop $k_{B}$ 10 iterations.  The procedure described above follows a first-order gradient-based meta-learning algorithm and its described in Algorithm \ref{alg:main}.

\begin{algorithm}[t]
 \caption{Training of the proposed architecture model. All conducted experiments in the paper used the default values: $n_{\mathrm{iter}} = 200,000$, $\alpha_{\mathrm{disc}} = \alpha_{\mathrm{gen}} = 0.0001$, $m = 16$, $k_{A}=1$, $k_{B}=10$, $n_{\mathrm{gen}}=5.$ }
 \label{alg:main}
 \begin{algorithmic}[1]
  \STATE Require: $n_{\mathrm{iter}}$, number of iterations. $\alpha$'s, learning rate. $m$, batch size. $n_{\mathrm{gen}}$, $k_{A}$, number of iterations inner-loop. $k_{B}$, number of iterations outer-loop. Number of skipped iterations of the generator per discriminator iteration.

  \STATE Require: $\phi_{0}$, initial generator and discriminator parameters.
  \STATE Initialize $\phi_{B} = \phi_{0}$
  \FOR {$i < n_{\mathrm{iter}}$}
  
  \STATE Sample a batch of images $\{\mathrm{\mathbf{x}}^{(z)}\}_{j=0}^m$
  \STATE Sample a task $\tau$
  \STATE Initialize $\phi_{A} = \phi_{B}$
  
  \FOR {$j < k_{A}$}
  
  \STATE $\#$ Train discriminator $D_{A}$
  \STATE $\phi_{A,\mathrm{disc}} \leftarrow \phi_{A,\mathrm{disc}} + \alpha_{\mathrm{disc}} \nabla_{\phi_{A}} \{ \mathcal{L}_{\mathrm{disc}}(\mathrm{\mathbf{x}},\mathrm{G_{A}(\mathrm{\mathbf{x}}})\}$
  \STATE $\#$ Train generator $G_{A}$
  \IF {$mod(i,n_{\mathrm{gen}}) = 0$}
  \STATE $\phi_{A,\mathrm{gen}} \leftarrow \phi_{A,\mathrm{gen}} + \alpha_{\mathrm{gen}} \nabla_{\phi_{A}} \{ \mathcal{L}_{\mathrm{gen}}(\mathrm{\mathbf{x}})\}$
  \ENDIF
  \ENDFOR
  
  \STATE Set gradients of $D_{B}$ and $G_{B}$ to $\phi_{0}-\phi_{A}$
  \STATE Perform step on $D_{B}$ and $G_{B}$ optimizers 
  \ENDFOR
  
  \STATE $\#$ Train few-shot
  \FOR {$j < k_{B}$}
  \STATE $\#$ Train discriminator $D_{B}$
  \STATE $\phi_{B,\mathrm{disc}} \leftarrow \phi_{B,\mathrm{disc}} + \alpha_{\mathrm{disc}} \nabla_{\phi_{B}} \{ \mathcal{L}_{\mathrm{disc}}(\mathrm{\mathbf{x}},\mathrm{G_{B}(\mathrm{\mathbf{x}}})\}$
  \STATE $\#$ Train generator $G_{B}$
  \IF {$mod(j,n_{\mathrm{gen}}) = 0$}
  \STATE $\phi_{B,\mathrm{gen}} \leftarrow \phi_{B,\mathrm{gen}} + \alpha_{\mathrm{gen}} \nabla_{\phi_{B}} \{ \mathcal{L}_{\mathrm{gen}}(\mathrm{\mathbf{x}})\}$
  \ENDIF
  \ENDFOR 
  \end{algorithmic} 
\end{algorithm}

\begin{figure*}[t]
  \begin{subfigure}[b]{0.5\linewidth}
    \centering
    \includegraphics[width=0.9\linewidth]{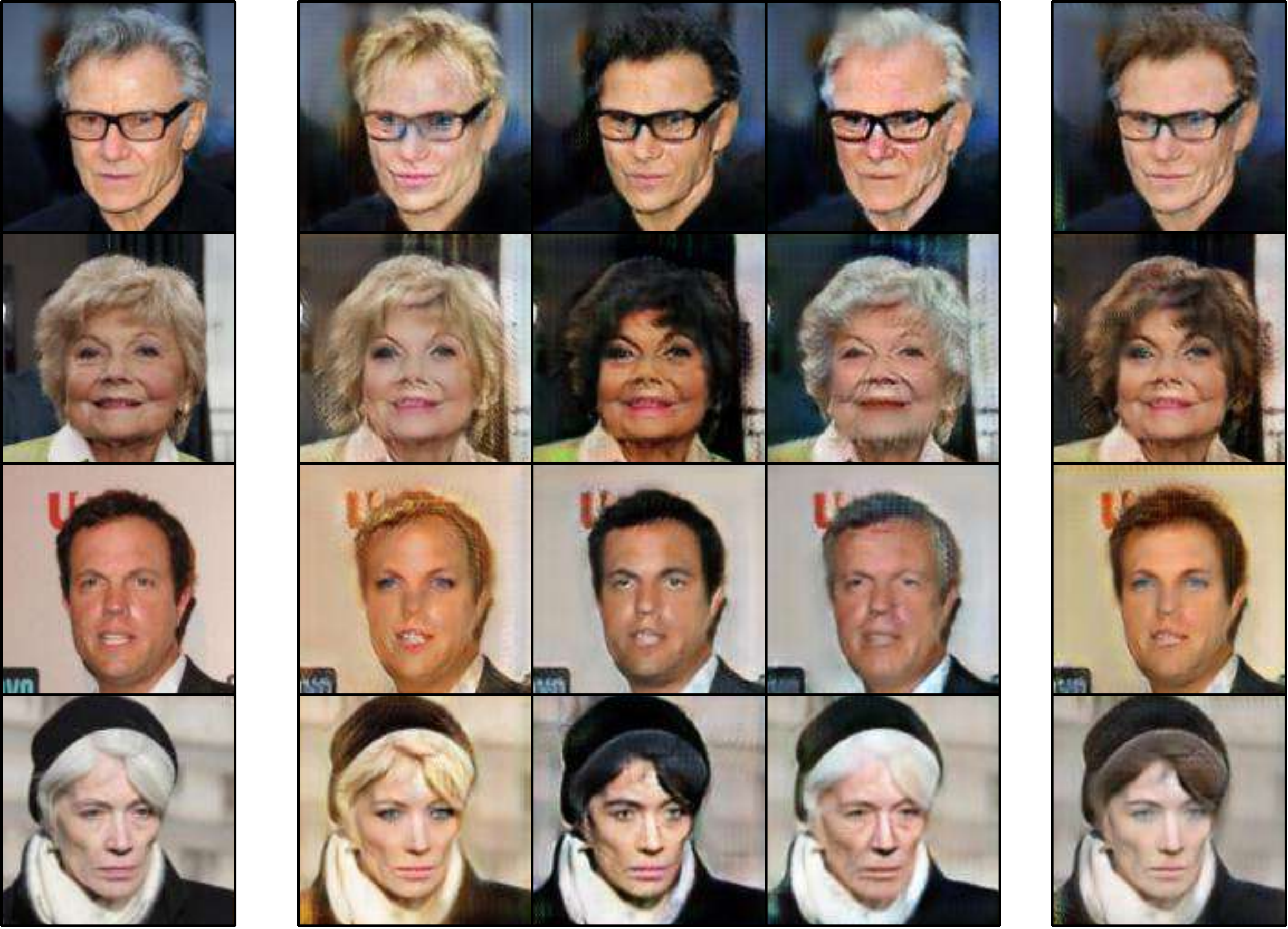} 
    \caption{Brown hair attribute transform.} 
    \label{fig:brown} 
    \vspace{2ex}
  \end{subfigure}
  \begin{subfigure}[b]{0.5\linewidth}
    \centering
    \includegraphics[width=0.9\linewidth]{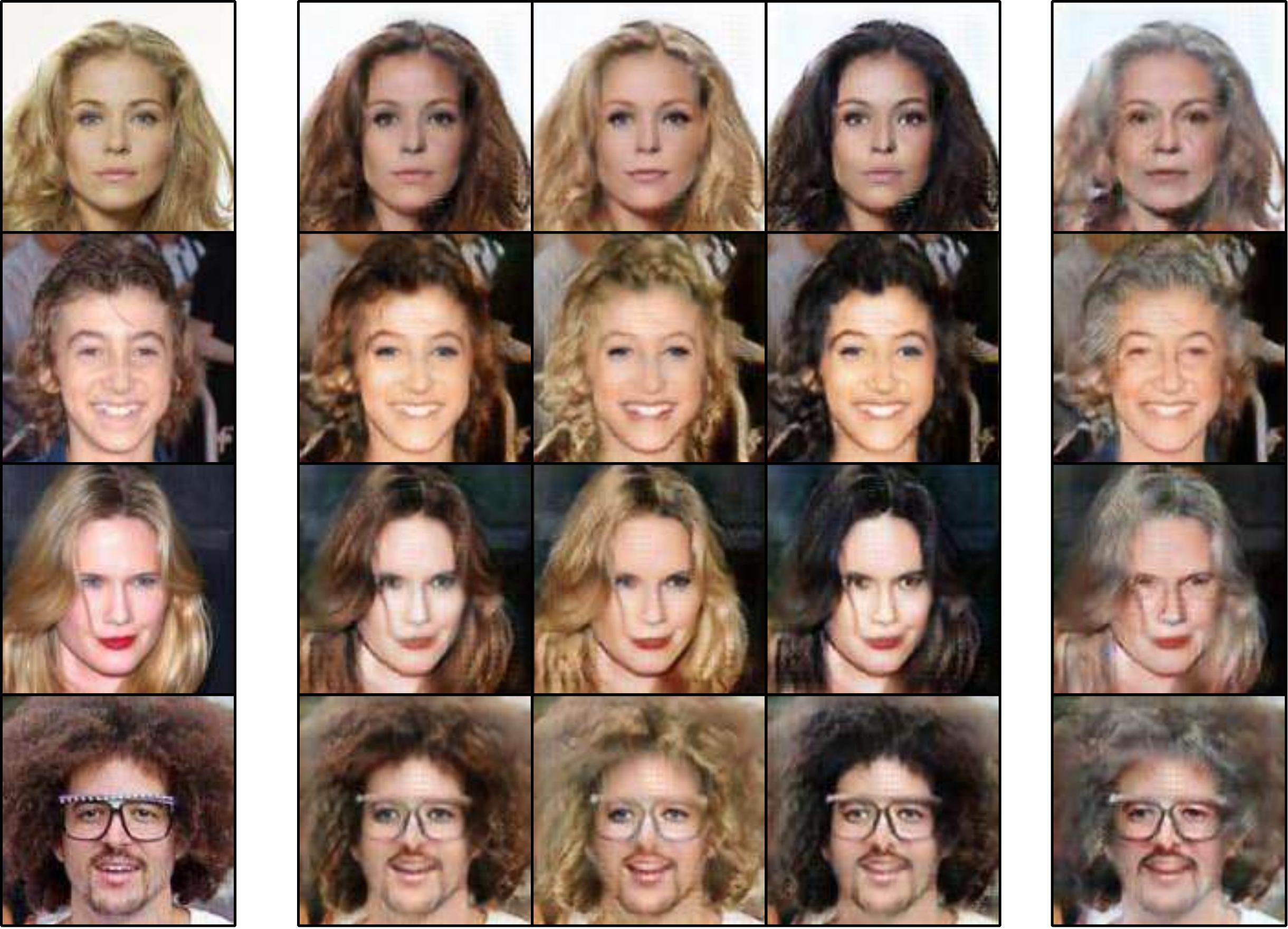} 
    \caption{Gray hair attribute transform.} 
    \label{fig:gray} 
    \vspace{2ex}
  \end{subfigure} 
  \begin{subfigure}[b]{0.5\linewidth}
    \centering
    \includegraphics[width=0.9\linewidth]{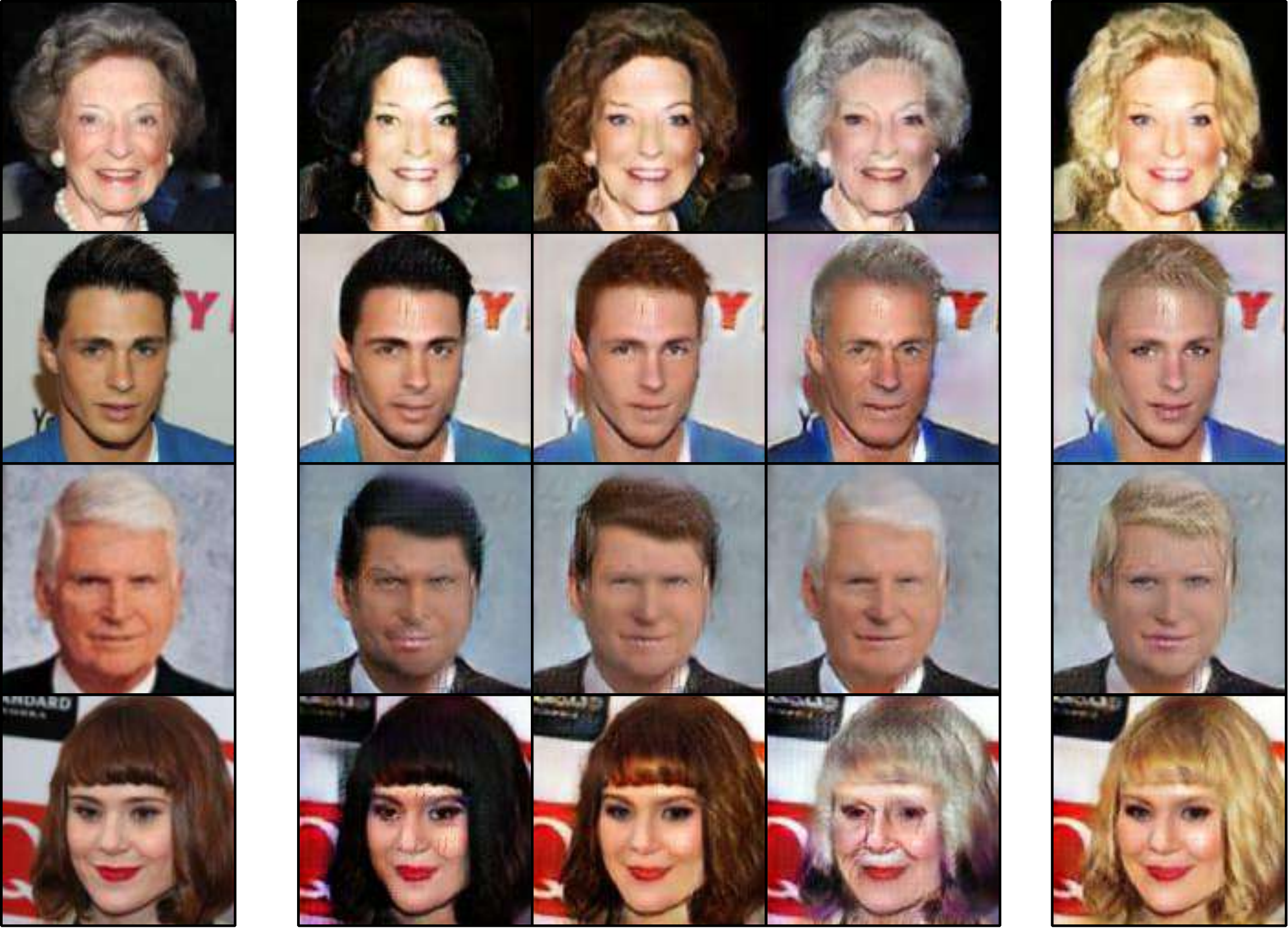} 
    \caption{Blond hair attribute transform.} 
    \label{fig:blond} 
  \end{subfigure}
  \begin{subfigure}[b]{0.5\linewidth}
    \centering
    \includegraphics[width=0.9\linewidth]{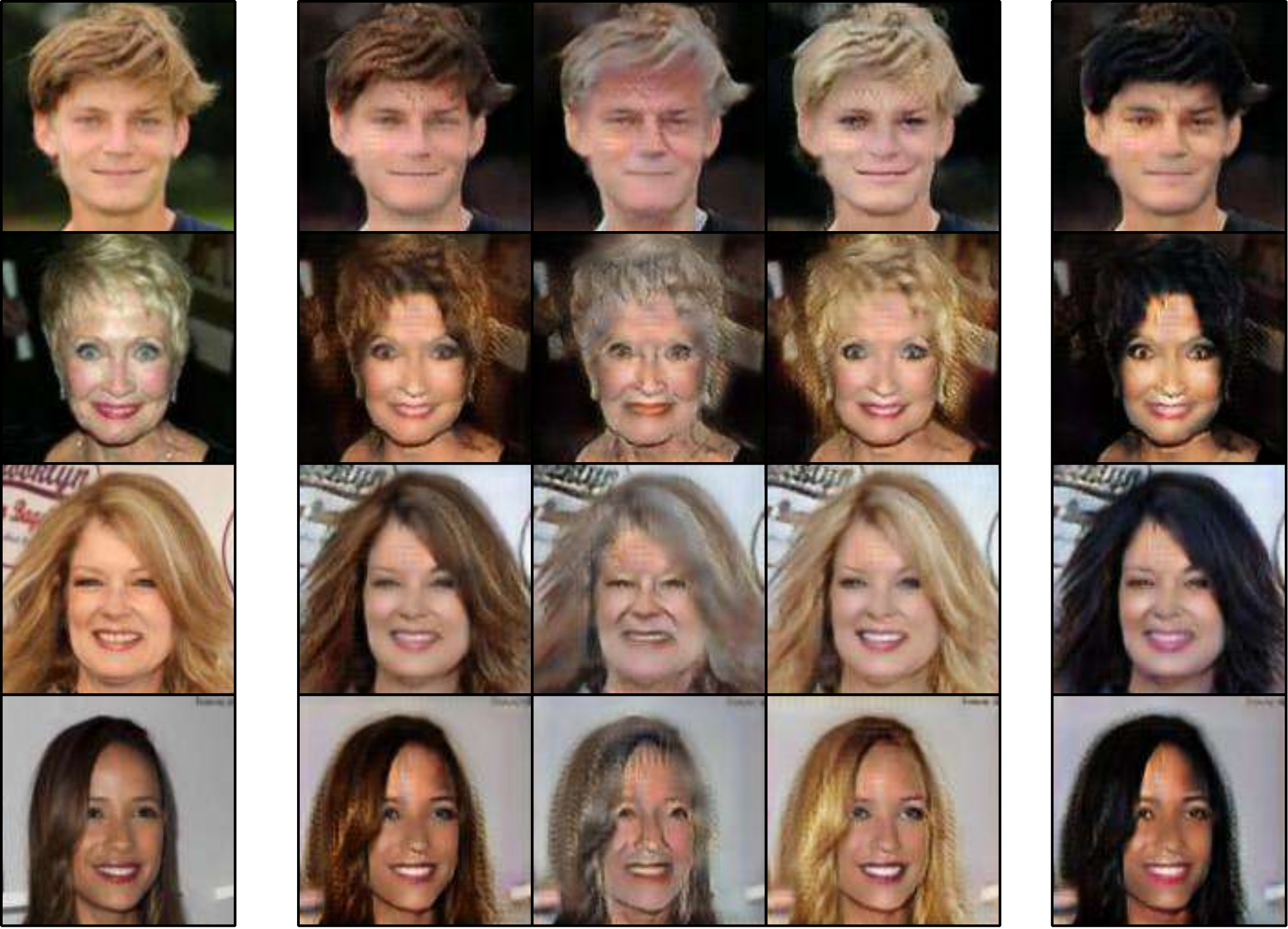} 
    \caption{Black hair attribute transform.} 
    \label{fig:black} 
  \end{subfigure} 
  \caption{Illustration of various few-shot hair attribute transfer testing results. Each sub-figure is an independent experiment, where the first column is the original image. The following three columns are the attribute learnt during the training and the last column is the target attribute which has been trained within a few-shot. }
  \label{fig:3} 
\end{figure*}

\subsection{Empirical Validation}

In this subsection, we present an empirical study of the results of our proposed method. We validate, that the attribute in the generated image, continuously changes with the coding vector $\tau$ (task). This phenomenon is known as attribute transfer or morphing.  In particular, we focus on hair color attributes: blond, black, brown and gray hair.

Figure \ref{fig:3} depicts four different experiments for hair attribute transfer. For example, Figure \ref{fig:brown} shows the result of training the images using the tasks blond, black and gray, during the normal training and then apply few-shot learning for the task brown hair. By implementing this procedure, the algorithm can find really suitable initialization weights, so that, when we fine-tune for the new task (e.g. brown hair) just with a few samples and a few iterations, the algorithm already converges towards a good
local minima. Additionally, the algorithm keeps the ability of transfer the attributes for which is originally trained (e.g. blond, black and gray hair).

By judging the results, we can conduct a qualitative evaluation that suggests a good behaviour of the model. It clearly shows the ability of generating natural-looking faces after applying the image-to-image transformation. Furthermore, it is important to notice that the gray hair attribute incorporates more information than just the color of the hair. We can observe that results which this attribute also look older. The reason for such an event is the entanglement of some attributes. In this dataset, the attribute gray hair is almost always related with old people. Therefore, when the model is pushed to learn the attribute, it cannot decouple the attributes old and gray hair.

\begin{figure*}[ht]
\centering
  \begin{subfigure}[b]{0.45\linewidth}
    \centering
    \includegraphics[width=0.95\linewidth]{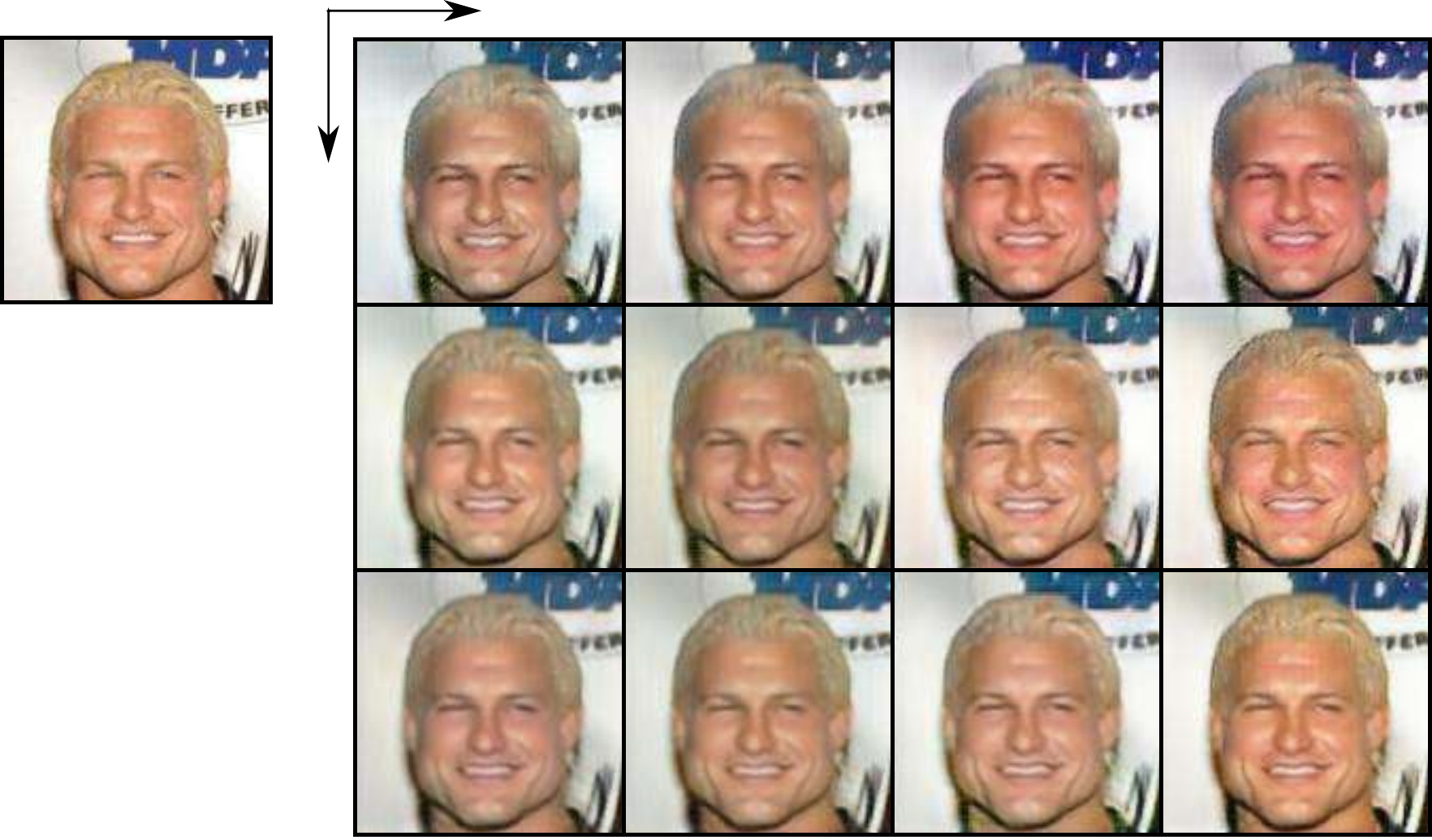}
    \caption{Blond to gray transformation.} 
    \label{fig:a} 
    \vspace{1ex}
  \end{subfigure}
  \begin{subfigure}[b]{0.45\linewidth}
    \centering
    \includegraphics[width=0.95\linewidth]{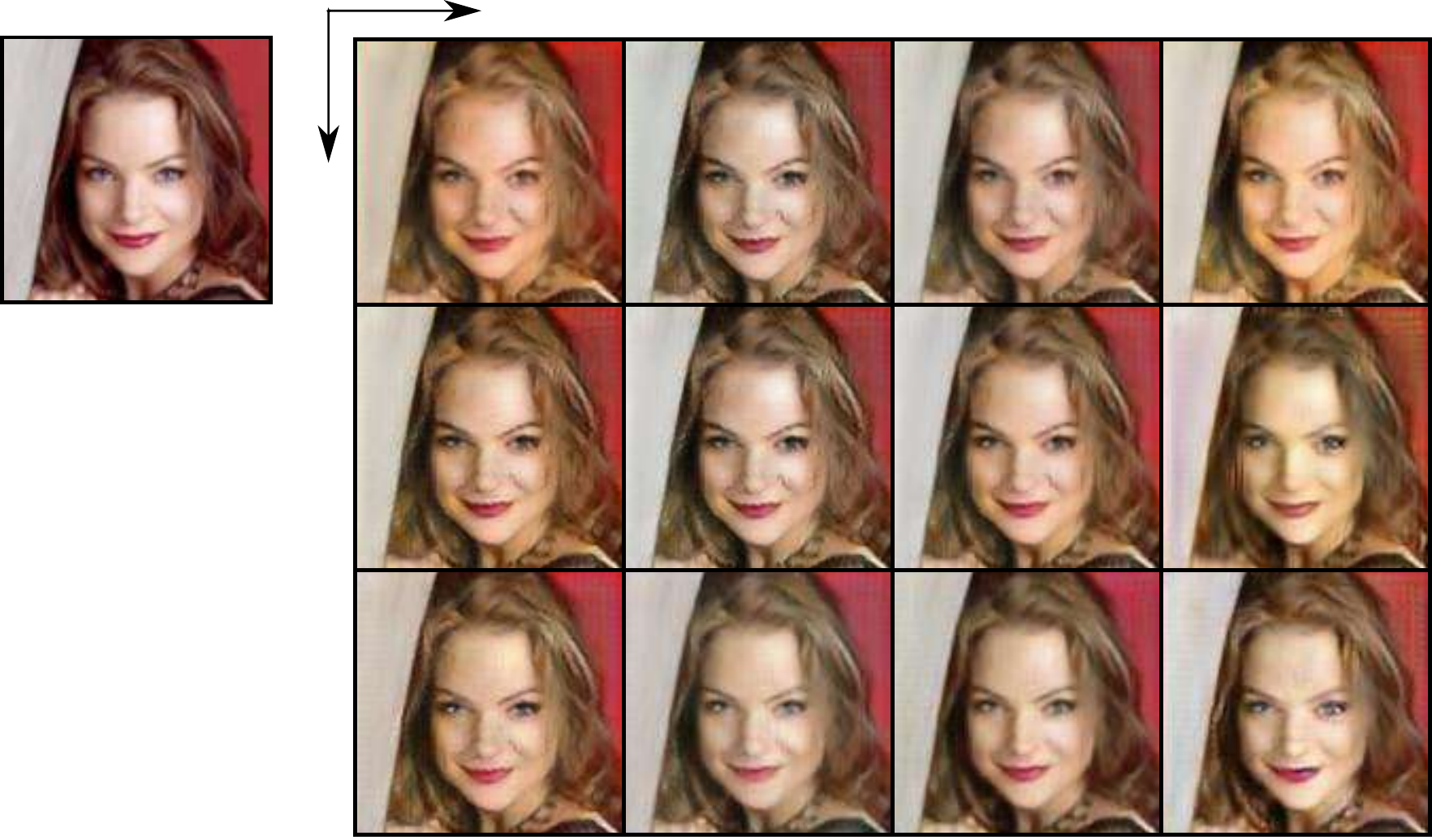}
    \caption{Brown to blond transformation.} 
    \label{fig:b} 
    \vspace{1ex}
  \end{subfigure} 
  \begin{subfigure}[b]{0.45\linewidth}
    \centering
    \includegraphics[width=0.95\linewidth]{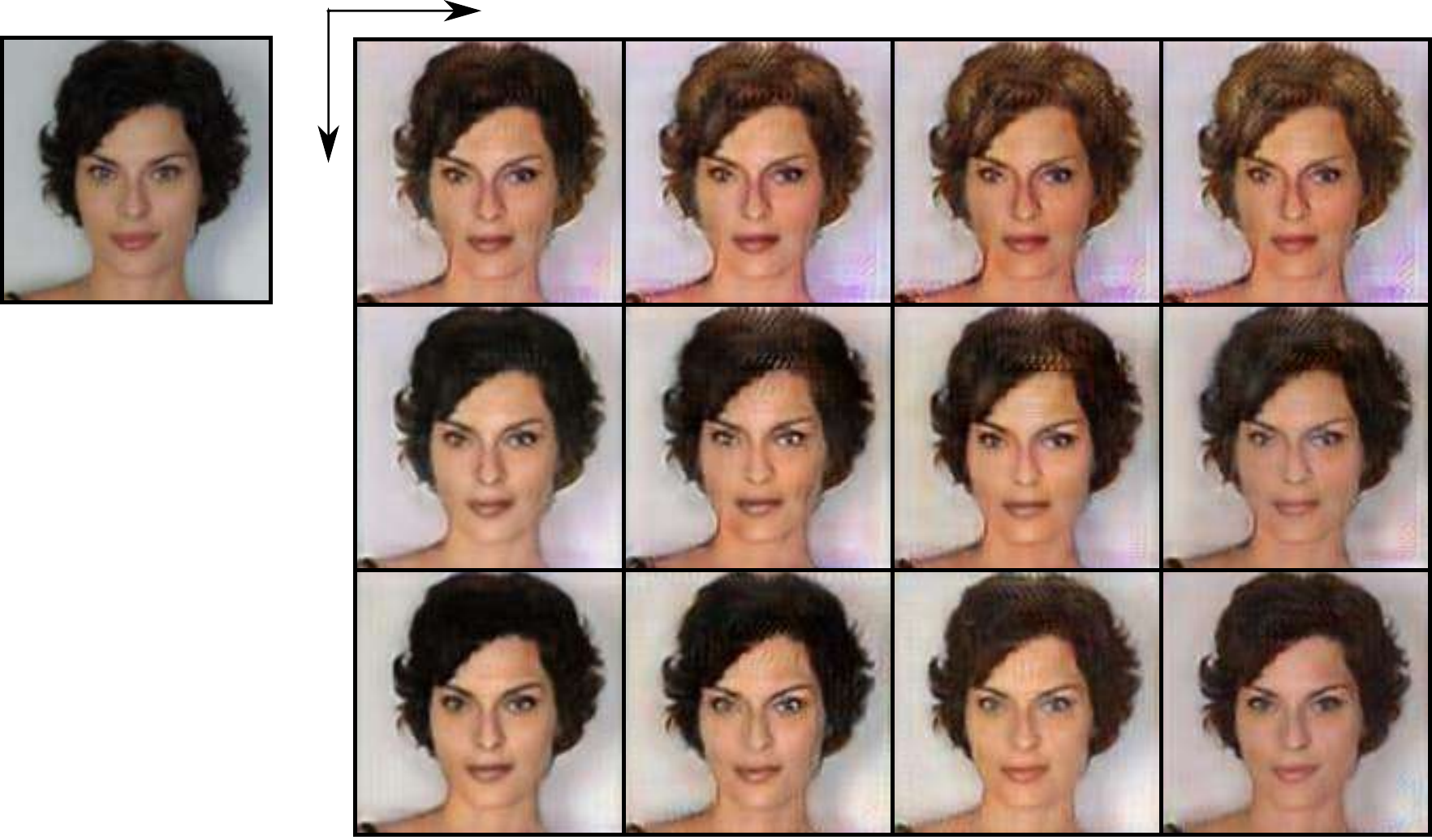}
    \caption{Black to brown transformation.}  
    \label{fig:c} 
  \end{subfigure}
  \begin{subfigure}[b]{0.45\linewidth}
    \centering
    \includegraphics[width=0.95\linewidth]{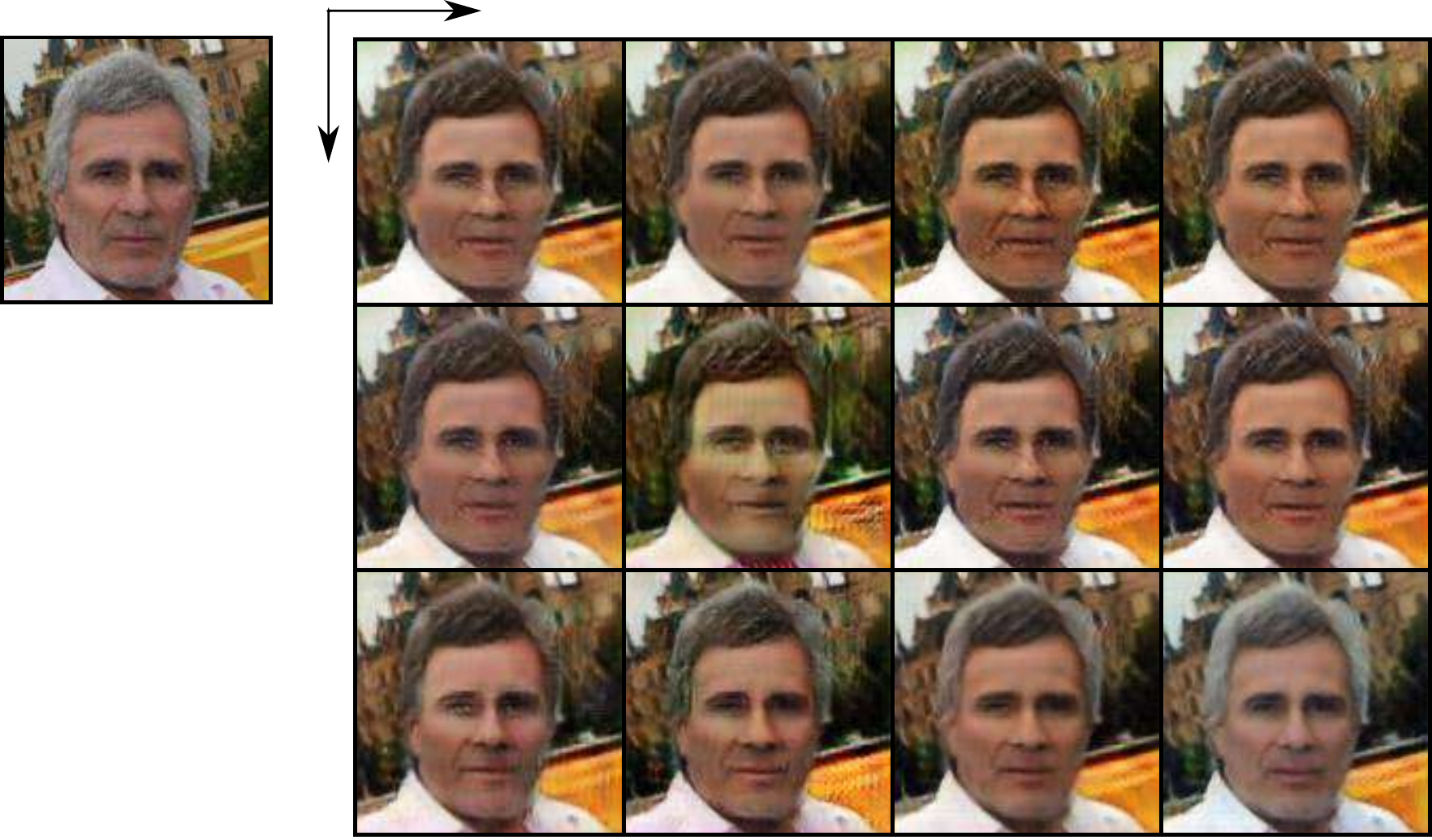}
    \caption{Gray to black transformation.} 
    \label{fig:d} 
  \end{subfigure} 
    \caption{Illustration of various few-shot hair attribute transfer testing results. Each sub-figure is an independent experiment, where we have evaluated the grid search space made of number of samples used for fine-tuning (4,8,16 or 32 column-wise from left to right), and number of gradient updates (10, 100, or 1000 row-size from top to bottom). We can see that best results are always found when we used 32 samples and 10 gradient step updates (upper-right corner).}
\label{fig:4}
\end{figure*}

\subsection{Ablation Study}

We present further experiments that support the proposed few-shot algorithm and try to build a general intuition about how sensitive our model is. We conduct a large set of tests for several hair attributes, where we modify the amount of samples of the unseen target class (4,8,16 or 32 samples) and the number of gradient steps (10, 100 or 1000) building in this way, a grid search space. Figure \ref{fig:4} shows four examples and the results after applying the different configuration setup. As expected, the more samples are used for fine-tuning towards the target task, the better the results are. Therefore, experiments where 32 samples are used, offer in general the best performances. Surprisingly, the number of gradient updates shows a counter-intuitive results. The more we optimize towards the class, the worse results. This phenomenon can be explained by a generalization intuition. Given a loss landscape, after 10 iterations we might find in a wide local minima, however, if we keep on optimizing it is likely that we end up in a much sharper minima where at testing time will yield much worse results. This is what we can observe, the more gradient updates, the less attribute transfer is present on the images.\\

\section{Discussion and Conclusion}

We have shown that meta-learning can be used to effectively train generative models for few-shot attribute transfer. Using these techniques on attributes, we can learn to generate  images  containing unseen attributes with just a few samples. This is done with no lengthy inference time, no external memory and no additional  data. Results show that our approach is able to learn and generate attribute given complex image structures like faces. The low amount of data required to generate images, once the model is pre-trained, opens the door to several applications that were previously gated by the high amount of data required. We see many interesting avenues of future work including combining different types of attribute transform such as hair color with smiling, eyeglasses, and other facial attributes.

{\small
\bibliographystyle{abbrv}
\bibliography{bibliography}

\begin{thebibliography}{10}

\bibitem{antoniou2018train}
A.~Antoniou, H.~Edwards, and A.~Storkey.
\newblock How to train your maml.
\newblock {\em arXiv preprint arXiv:1810.09502}, 2018.

\bibitem{bansal2018recycle}
A.~Bansal, S.~Ma, D.~Ramanan, and Y.~Sheikh.
\newblock Recycle-gan: Unsupervised video retargeting.
\newblock In {\em Proceedings of the European Conference on Computer Vision
  (ECCV)}, pages 119--135, 2018.

\bibitem{bao2017cvae}
J.~Bao, D.~Chen, F.~Wen, H.~Li, and G.~Hua.
\newblock Cvae-gan: fine-grained image generation through asymmetric training.
\newblock In {\em Proceedings of the IEEE International Conference on Computer
  Vision}, pages 2745--2754, 2017.

\bibitem{bartunov2018few}
S.~Bartunov and D.~Vetrov.
\newblock Few-shot generative modelling with generative matching networks.
\newblock In {\em International Conference on Artificial Intelligence and
  Statistics}, pages 670--678, 2018.

\bibitem{choi2018stargan}
Y.~Choi, M.~Choi, M.~Kim, J.-W. Ha, S.~Kim, and J.~Choo.
\newblock Stargan: Unified generative adversarial networks for multi-domain
  image-to-image translation.
\newblock In {\em Proceedings of the IEEE Conference on Computer Vision and
  Pattern Recognition}, pages 8789--8797, 2018.

\bibitem{chongxuan2017triple}
L.~Chongxuan, T.~Xu, J.~Zhu, and B.~Zhang.
\newblock Triple generative adversarial nets.
\newblock In {\em Advances in neural information processing systems}, pages
  4088--4098, 2017.

\bibitem{clouatre2019figr}
L.~Clou{\^a}tre and M.~Demers.
\newblock Figr: Few-shot image generation with reptile.
\newblock {\em arXiv preprint arXiv:1901.02199}, 2019.

\bibitem{creswell2017adversarial}
A.~Creswell, Y.~Mohamied, B.~Sengupta, and A.~A. Bharath.
\newblock Adversarial information factorization.
\newblock {\em arXiv preprint arXiv:1711.05175}, 2017.

\bibitem{dodge2017study}
S.~Dodge and L.~Karam.
\newblock A study and comparison of human and deep learning recognition
  performance under visual distortions.
\newblock In {\em 2017 26th international conference on computer communication
  and networks (ICCCN)}, pages 1--7. IEEE, 2017.

\bibitem{finn2017model}
C.~Finn, P.~Abbeel, and S.~Levine.
\newblock Model-agnostic meta-learning for fast adaptation of deep networks.
\newblock In {\em Proceedings of the 34th International Conference on Machine
  Learning-Volume 70}, pages 1126--1135. JMLR. org, 2017.

\bibitem{finn2018probabilistic}
C.~Finn, K.~Xu, and S.~Levine.
\newblock Probabilistic model-agnostic meta-learning.
\newblock In {\em Advances in Neural Information Processing Systems}, pages
  9516--9527, 2018.

\bibitem{goodfellow2014generative}
I.~Goodfellow, J.~Pouget-Abadie, M.~Mirza, B.~Xu, D.~Warde-Farley, S.~Ozair,
  A.~Courville, and Y.~Bengio.
\newblock Generative adversarial nets.
\newblock In {\em Advances in neural information processing systems}, pages
  2672--2680, 2014.

\bibitem{gulrajani2017improved}
I.~Gulrajani, F.~Ahmed, M.~Arjovsky, V.~Dumoulin, and A.~C. Courville.
\newblock Improved training of wasserstein gans.
\newblock In {\em Advances in Neural Information Processing Systems}, pages
  5767--5777, 2017.

\bibitem{huang2018multimodal}
X.~Huang, M.-Y. Liu, S.~Belongie, and J.~Kautz.
\newblock Multimodal unsupervised image-to-image translation.
\newblock In {\em Proceedings of the European Conference on Computer Vision
  (ECCV)}, pages 172--189, 2018.

\bibitem{iizuka2017globally}
S.~Iizuka, E.~Simo-Serra, and H.~Ishikawa.
\newblock Globally and locally consistent image completion.
\newblock {\em ACM Transactions on Graphics (ToG)}, 36(4):107, 2017.

\bibitem{isola2017image}
P.~Isola, J.-Y. Zhu, T.~Zhou, and A.~A. Efros.
\newblock Image-to-image translation with conditional adversarial networks.
\newblock {\em arXiv preprint}, 2017.

\bibitem{karras2017progressive}
T.~Karras, T.~Aila, S.~Laine, and J.~Lehtinen.
\newblock Progressive growing of gans for improved quality, stability, and
  variation.
\newblock {\em arXiv preprint arXiv:1710.10196}, 2017.

\bibitem{kim2017learning}
T.~Kim, M.~Cha, H.~Kim, J.~K. Lee, and J.~Kim.
\newblock Learning to discover cross-domain relations with generative
  adversarial networks.
\newblock In {\em Proceedings of the 34th International Conference on Machine
  Learning-Volume 70}, pages 1857--1865. JMLR. org, 2017.

\bibitem{kingma2014adam}
D.~P. Kingma and J.~Ba.
\newblock Adam: A method for stochastic optimization.
\newblock {\em arXiv preprint arXiv:1412.6980}, 2014.

\bibitem{koch2015siamese}
G.~Koch, R.~Zemel, and R.~Salakhutdinov.
\newblock Siamese neural networks for one-shot image recognition.
\newblock In {\em ICML deep learning workshop}, volume~2, 2015.

\bibitem{lake2011one}
B.~Lake, R.~Salakhutdinov, J.~Gross, and J.~Tenenbaum.
\newblock One shot learning of simple visual concepts.
\newblock In {\em Proceedings of the Annual Meeting of the Cognitive Science
  Society}, volume~33, 2011.

\bibitem{lecun2010mnist}
Y.~LeCun, C.~Cortes, and C.~Burges.
\newblock Mnist handwritten digit database.
\newblock {\em AT\&T Labs [Online]. Available: http://yann. lecun.
  com/exdb/mnist}, 2:18, 2010.

\bibitem{ledig2017photo}
C.~Ledig, L.~Theis, F.~Husz{\'a}r, J.~Caballero, A.~Cunningham, A.~Acosta,
  A.~Aitken, A.~Tejani, J.~Totz, Z.~Wang, et~al.
\newblock Photo-realistic single image super-resolution using a generative
  adversarial network.
\newblock In {\em Proceedings of the IEEE conference on computer vision and
  pattern recognition}, pages 4681--4690, 2017.

\bibitem{li2017generative}
Y.~Li, S.~Liu, J.~Yang, and M.-H. Yang.
\newblock Generative face completion.
\newblock In {\em Proceedings of the IEEE Conference on Computer Vision and
  Pattern Recognition}, pages 3911--3919, 2017.

\bibitem{liao2017visual}
J.~Liao, Y.~Yao, L.~Yuan, G.~Hua, and S.~B. Kang.
\newblock Visual attribute transfer through deep image analogy.
\newblock {\em arXiv preprint arXiv:1705.01088}, 2017.

\bibitem{liu2015deep}
Z.~Liu, P.~Luo, X.~Wang, and X.~Tang.
\newblock Deep learning face attributes in the wild.
\newblock In {\em Proceedings of the IEEE international conference on computer
  vision}, pages 3730--3738, 2015.

\bibitem{mo2018instanceaware}
S.~Mo, M.~Cho, and J.~Shin.
\newblock Instance-aware image-to-image translation.
\newblock In {\em International Conference on Learning Representations}, 2019.

\bibitem{munkhdalai2017meta}
T.~Munkhdalai and H.~Yu.
\newblock Meta networks.
\newblock In {\em Proceedings of the 34th International Conference on Machine
  Learning-Volume 70}, pages 2554--2563. JMLR. org, 2017.

\bibitem{nichol2018first}
A.~Nichol, J.~Achiam, and J.~Schulman.
\newblock On first-order meta-learning algorithms.
\newblock {\em arXiv preprint arXiv:1803.02999}, 2018.

\bibitem{pathak2016context}
D.~Pathak, P.~Krahenbuhl, J.~Donahue, T.~Darrell, and A.~A. Efros.
\newblock Context encoders: Feature learning by inpainting.
\newblock In {\em Proceedings of the IEEE conference on computer vision and
  pattern recognition}, pages 2536--2544, 2016.

\bibitem{ravi2016optimization}
S.~Ravi and H.~Larochelle.
\newblock Optimization as a model for few-shot learning.
\newblock 2016.

\bibitem{rezende2016one}
D.~J. Rezende, S.~Mohamed, I.~Danihelka, K.~Gregor, and D.~Wierstra.
\newblock One-shot generalization in deep generative models.
\newblock {\em arXiv preprint arXiv:1603.05106}, 2016.

\bibitem{santoro2016meta}
A.~Santoro, S.~Bartunov, M.~Botvinick, D.~Wierstra, and T.~Lillicrap.
\newblock Meta-learning with memory-augmented neural networks.
\newblock In {\em International conference on machine learning}, pages
  1842--1850, 2016.

\bibitem{shen2017style}
T.~Shen, T.~Lei, R.~Barzilay, and T.~Jaakkola.
\newblock Style transfer from non-parallel text by cross-alignment.
\newblock In {\em Advances in neural information processing systems}, pages
  6830--6841, 2017.

\bibitem{snell2017prototypical}
J.~Snell, K.~Swersky, and R.~Zemel.
\newblock Prototypical networks for few-shot learning.
\newblock In {\em Advances in Neural Information Processing Systems}, pages
  4077--4087, 2017.

\bibitem{sung2018learning}
F.~Sung, Y.~Yang, L.~Zhang, T.~Xiang, P.~H. Torr, and T.~M. Hospedales.
\newblock Learning to compare: Relation network for few-shot learning.
\newblock In {\em Proceedings of the IEEE Conference on Computer Vision and
  Pattern Recognition}, pages 1199--1208, 2018.

\bibitem{vinyals2016matching}
O.~Vinyals, C.~Blundell, T.~Lillicrap, D.~Wierstra, et~al.
\newblock Matching networks for one shot learning.
\newblock In {\em Advances in neural information processing systems}, pages
  3630--3638, 2016.

\bibitem{wang2018video}
T.-C. Wang, M.-Y. Liu, J.-Y. Zhu, G.~Liu, A.~Tao, J.~Kautz, and B.~Catanzaro.
\newblock Video-to-video synthesis.
\newblock {\em arXiv preprint arXiv:1808.06601}, 2018.

\bibitem{yeh2017semantic}
R.~A. Yeh, C.~Chen, T.~Yian~Lim, A.~G. Schwing, M.~Hasegawa-Johnson, and M.~N.
  Do.
\newblock Semantic image inpainting with deep generative models.
\newblock In {\em Proceedings of the IEEE Conference on Computer Vision and
  Pattern Recognition}, pages 5485--5493, 2017.

\bibitem{yu2018generative}
J.~Yu, Z.~Lin, J.~Yang, X.~Shen, X.~Lu, and T.~S. Huang.
\newblock Generative image inpainting with contextual attention.
\newblock In {\em Proceedings of the IEEE Conference on Computer Vision and
  Pattern Recognition}, pages 5505--5514, 2018.

\bibitem{zhang2017stackgan}
H.~Zhang, T.~Xu, H.~Li, S.~Zhang, X.~Wang, X.~Huang, and D.~N. Metaxas.
\newblock Stackgan: Text to photo-realistic image synthesis with stacked
  generative adversarial networks.
\newblock In {\em Proceedings of the IEEE International Conference on Computer
  Vision}, pages 5907--5915, 2017.

\bibitem{zhang2016colorful}
R.~Zhang, P.~Isola, and A.~A. Efros.
\newblock Colorful image colorization.
\newblock In {\em European conference on computer vision}, pages 649--666.
  Springer, 2016.

\bibitem{zhu2017unpaired}
J.-Y. Zhu, T.~Park, P.~Isola, and A.~A. Efros.
\newblock Unpaired image-to-image translation using cycle-consistent
  adversarial networks.
\newblock {\em arXiv preprint}, 2017.

\bibitem{zhu2017toward}
J.-Y. Zhu, R.~Zhang, D.~Pathak, T.~Darrell, A.~A. Efros, O.~Wang, and
  E.~Shechtman.
\newblock Toward multimodal image-to-image translation.
\newblock In {\em Advances in Neural Information Processing Systems}, pages
  465--476, 2017.

\end{thebibliography}
}

\end{document}